# Face Recognition: A Novel Multi-Level Taxonomy based Survey

Alireza Sepas-Moghaddam [1,*], Fernando Pereira [1], Paulo Lobato Correia [1]

[1] Instituto de Telecomunicações, Instituto Superior Técnico – Universidade de Lisboa, , Lisbon, Portugal
[*] alireza@lx.it.pt

**Abstract:** In a world where security issues have been gaining growing importance, face recognition systems have attracted increasing attention in multiple application areas, ranging from forensics and surveillance to commerce and entertainment. To help understanding the landscape and abstraction levels relevant for face recognition systems, face recognition taxonomies allow a deeper dissection and comparison of the existing solutions. This paper proposes a new, more encompassing and richer multi-level face recognition taxonomy, facilitating the organization and categorization of available and emerging face recognition solutions; this taxonomy may also guide researchers in the development of more efficient face recognition solutions. The proposed multi-level taxonomy considers levels related to the face structure, feature support and feature extraction approach. Following the proposed taxonomy, a comprehensive survey of representative face recognition solutions is presented. The paper concludes with a discussion on current algorithmic and application related challenges which may define future research directions for face recognition.

## 1. Introduction

Face recognition systems have been successfully used in multiple application areas with high acceptability, collectability and universality [1] [2]. After the first automatic face recognition algorithms emerged more than four decades ago [3], this field has attracted much research and witnesses incredible progress, with a very large number of face recognition solutions being used in multiple application areas. Face recognition technology is assuming an increasingly important role in our everyday life, what also brings ethical and privacy dilemmas about how the captured facial information and the corresponding identity, as a special category of personal data, should be used, stored and shared.

According to [4], "taxonomy is the practice and science of classification of things or concepts, including the principles that underlie such classification". The availability of a taxonomy in a certain field allows to organize/classify/abstract the 'things' (in this case, the face recognition solutions) with two main benefits: i) regarding the present, it makes it easier to discuss and analyse the 'things' and abstract the deeper relations between them, thus providing a deeper knowledge and comprehension of the full landscape, notably in terms of strengths and weaknesses; ii) regarding the future, it makes it easier to understand the most promising research directions and their implications as the 'things' (in this case, the face recognition solutions) will not be isolated 'things' but rather 'things' in a taxonomical network, inheriting features, strengths and weaknesses from their taxonomy parents and peers.

Compiling a comprehensive survey of the available face recognition solutions is a challenging task, notably given the large number and diversity of solutions developed in the last decades. To help understanding the structure and abstraction levels that may be considered in face recognition solutions, a number of face recognition taxonomies have been proposed so far [5] [6] [7] [8] [9] [10] [11] [12] [13] [14]. Since the available face recognition taxonomies ignore some relevant levels of abstraction, which may be helpful for a more complete characterization of the face recognition landscape, this paper proposes a new, more encompassing and richer face recognition multi-level taxonomy.

The new taxonomy can be used to better understand the technological landscape in the area, facilitating the characterization and organization of available solutions, and guiding researchers in the development of more efficient face recognition solutions for given applications. The proposed multi-level taxonomy considers four levels, notably face structure, feature support, feature extraction approach, and feature extraction sub-approach. This paper also surveys representative state-of-the-art face recognition solutions according to the proposed multi-level taxonomy and discusses the current algorithmic and application related challenges and future research directions for face recognition systems.

The rest of the paper is organized as follows. Section 2 reviews the available face recognition taxonomies, to understand their benefits and limitations. Section 3 proposes a new, more encompassing and richer multi-level taxonomy for face recognition solutions. Section 4 surveys the state-of-the-art on face recognition under the umbrella of the proposed multi-level face recognition taxonomy and discusses the evolutional trends of face recognition over time. Finally, Section 5 discusses some face recognition challenges and identifies some relevant future research directions.

## 2. Reviewing existing face recognition taxonomies

Several face recognition taxonomies have been proposed in the literature [5] [6] [7] [8] [9] [10] [11] [12] [13] [14], as summarized in Table 1. This table includes information about the abstraction level(s) considered as well as the corresponding classes – notice that some taxonomies may use a different terminology. Excluding the taxonomy proposed in [10], all the other taxonomies listed in Table 1 have been developed based on a single abstraction level to organize face recognition solutions, thus proposing a specific taxonomical point of view.

**Table 1:** Summary of available face recognition taxonomies and corresponding abstraction levels and classes.

| Ref. | Year | Abstraction Level(s) | Classes |
|---|---|---|---|
| [5] | 2003 | Feature extraction | Appearance based; Feature based; Hybrid |
| [6] | 2006 | Feature extraction | Holistic based; Local based; Hybrid |
| [7] | 2009 | Sensing modality | 2D; Video; 3D; infra-red data |
| [8] | 2009 | Modality matching | 2D vs. 2D; 2D vs. 2D via 3D; 3D vs. 3D; Video vs. 3D; Multimodal 2D+3D |
| [9] | 2010 | Facial features | Soft features; Anthropometric features; Unstructured and micro-level features |
| [10] | 2011 | Pose dependency | Pose-dependant; Pose-invariant |
| [10] | 2011 | Feature extraction | Appearance based; Feature based; Hybrid |
| [11] | 2012 | Feature extraction | Holistic based; Feature based; Model based; Hybrid |
| [12] | 2014 | Sensing modality | 2D; 3D; Multimodal 2D+3D |
| [13] | 2016 | Illumination dependency | Illumination normalization; Illumination modeling; Illumination in variation |
| [14] | 2016 | Pose dependency | Pose-robust; Multi-view learning; Face synthesis; Hybrid |

The face recognition taxonomies proposed in [5], [6], and [11] consider as abstraction level the type of feature extraction. These taxonomies include a 'hybrid' class to cover solutions combining two or more feature extraction classes to boost the face recognition performance. The taxonomy proposed in [9] is also related to the features used for recognition, but relies on the type of facial features and not on the feature extraction approach. This taxonomy considers three classes, notably: i) soft features, such as gender, race, and age, encompassing the global nature of the face, which are mostly useful for indexing and reducing the search space; ii) anthropometric features, such as facial landmarks and their relations, thus taking the facial structures as the most discriminative features; and iii) unstructured and micro-level features, such as tattoos, moles and scars, which may facilitate the discrimination between *short-listed* subjects.

Two recent face recognition taxonomies organize the face recognition solutions depending on the specific conditions addressed, such as illumination [13] and pose [14] dependencies, not being encompassing enough to consider every face recognition solution. Other taxonomies organize the face recognition solutions based on the sensing modality [7] [12], or on the adopted modality matching, to consider the cases where the enrolment and test data are captured using different imaging modalities [8]. To summarize, the reviewed single-level taxonomies classify face recognition solutions based on a specific abstraction point of view, ignoring other abstraction levels, along with their relations.

A multi-level taxonomy, would be useful for a more complete characterization and organization of face recognition solutions. Accordingly, a multi-level taxonomy has been proposed in [10], adopting two abstraction levels to organize the face recognition solutions, notably pose-dependency and matching features (equivalent to feature extraction). However, this taxonomy still ignores some relevant abstraction levels, such as the face structure and feature support, which should be considered for a more complete characterization of face recognition solutions.

The major improvements of the proposed multi-level taxonomy proposed in this paper are:

1. The proposed taxonomy considers face structure and feature support levels for a more complete characterization of face recognition solutions. These levels provide critical information on the way face recognition solutions deal with the facial structure and what is the region of spatial support for feature extraction, impacting the selection of the corresponding approach.
2. The feature extraction approach level in the proposed taxonomy (equivalent to the matching features level in [10]) considers a richer and more complete set of classes, rather than only feature based and appearance based classes as in [10].
3. The feature extraction approach level is further divided into feature extraction sub-approaches to more deeply understand the technological landscape in the context of feature extraction.

Regarding pose-dependency, this level was considered not relevant enough as most recent face recognition solutions are pose-invariant.

In summary, the proposed multi-level taxonomy provides a more encompassing and richer taxonomy for face recognition as may, thus, bring additional value in understanding the face recognition landscape.

## 3. Proposing a novel multi-level face recognition taxonomy

Following the taxonomy limitations highlighted in Section 2, this paper proposes a novel, more encompassing and richer multi-level face recognition taxonomy, which can help to better understand the technological landscape in the area, thus facilitating the characterization, organization and comparison of face recognition solutions. The proposed multi-level face recognition taxonomy, illustrated in Figure 1, considers four taxonomy levels:

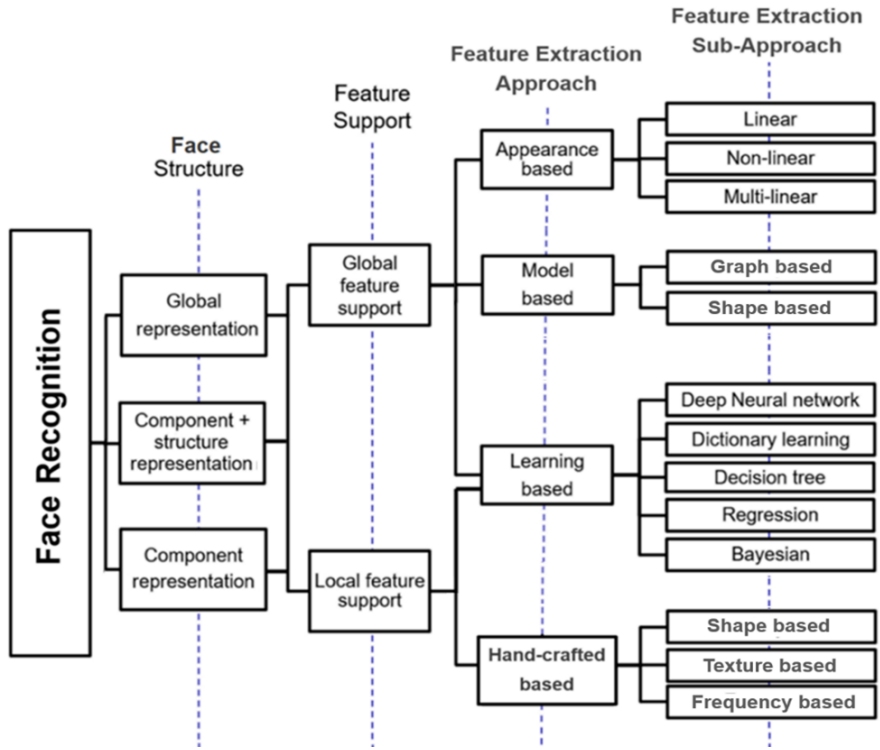

**Figure 1:** Proposed multi-level face recognition taxonomy.

1. **Face structure** – This level relates to the way a recognition solution deals with the facial structure, considering three classes: i) *global representation*, dealing with the face as a whole (see Figure 2.a); ii) *component + structure representation*, relying only on the characteristics of some face components, such as eyes, nose, mouth, etc., along with their relations (see Figure 2.b); and iii) *component representation*, dealing independently with a meaningful selection of face components, without any consideration of the relations between them (see Figure 2.c).

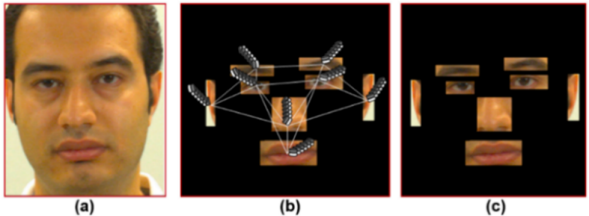

**Figure 2:** Face structure level: (a) global; (b) component + structure; and (c) component representation face structures.

2. **Feature support** – This level is related to the (spatial) region of support considered for feature extraction, which can be either *global* or *local*. Global feature support implies that all the selected facial structure area is considered as region of support for feature extraction, corresponding to either the full face (Figure 3.a) or a full face component (Figure 3.b), without any further partitioning. On the contrary, local feature support implies that the region of support for feature extraction is a smaller part of either the full face (Figure 3.c) or the face component (Figure 3.d). A local region of support can have different characteristics, for instance in terms of topology, size, overlapping, among others; a much used type of partitioning is the simple square based division of the face or component.

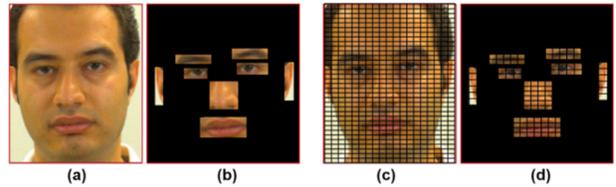

**Figure 3:** Feature support level: Global feature support with (a) global and (b) component face structures; Local feature support with (c) global and (d) component face structures. The square blocks in (c) and (d) represent the local (spatial) support considered for feature extraction.

3. **Feature extraction approach** – This level is related to the specific feature extraction approach, which may be classified as: i) *appearance* based, deriving features by using statistical transformations from the intensity data; ii) *model based*, deriving features based on geometrical characteristics of the face; iii) *learning based*, deriving features by modelling and learning relationships from the input data; and iv) *hand-crafted based*, deriving features from pre-selected elementary characteristics.

4. **Feature extraction sub-approach** – The last level considered in the proposed taxonomy is a sub-category of the previous one, to allow identifying the specific family of techniques used by the selected feature extraction approach.

Regarding the feature extraction approach and sub-approach levels, these are directly related to the specific feature description technology adopted, as discussed in the following.

*Appearance b*ased face recognition solutions map the input data into a lower dimensional space, while retaining the most relevant information. These solutions are generally sensitive to face variations, such as occlusions as well as scale, pose and expression changes, as they do not consider any specific knowledge about the face structure. Appearance based feature extraction solutions can be divided into: i) *linear* solutions, such as Principle Component Analysis (PCA) [15] and Independent Component Analysis (ICA) [16], performing an optimal linear mapping to a lower dimensional space to extract the representative features; ii) *non-linear* solutions, such as kernel PCA [17], exploiting the non-linear structure of face patterns to compute a non-linear mapping; and iii) *multi-linear*, such as generalized PCA [18], extracting information from high dimensional data while retaining its natural structure, thus providing more compact representations than linear solutions for high dimensional data.

*Model based* face recognition solutions derive features based on geometrical characteristics of the face. These solutions are generally less sensitive to face variations as they consider structural information from the face, thus requiring accurate landmark localization prior to feature extraction. Model based feature extraction solutions can be divided into: i) *graph based* solutions, such as Elastic Bunch Graph Matching (EBGM) [19], representing face features as a graph, where nodes store local information about face landmarks and edges represent relations, e.g. the distance between nodes, and a graph similarity function is used for matching; and ii) *shape based* solutions, such as the 3D Morphable Model (3DMM) [20], using landmarks to represent key face components, controlled by the model, and using shape similarity functions for matching.

*Learning based* solutions derive features by modelling and learning relationships from the input data. These solutions may offer robustness against facial variations, depending on the considered training data; however, they can be computationally more complex than solutions based on other feature extraction approaches, as they require initialization, training, and tuning of (hyper) parameters. In recent years, deep learning based solutions have been increasingly adopted for face recognition tasks and, not surprisingly, the current state-of-the-art on face recognition is dominated by deep neural networks, notably Convolutional Neural Networks (CNNs). Learning based face recognition solutions can be categorized into five families of techniques , including: i) *deep neural networks*, such as the VGG-Face descriptor [21], modelling the input data with high abstraction levels by using a deep graph with multiple processing layers to automatically learn features from the input data; ii) *dictionary learning* solutions, such as Kernel Extended Dictionary (KED) [22], finding a sparse input data feature representation in the form of a linear combination of basic elements; iii) *decision tree* solutions, such as Decision Pyramid (DP) [23], representing features as the result of a series of decisions; iv) *regression* solutions, such as Logistic Regression (LR) [24], iteratively refining the relation between variables using an error measure for the predictions made by the adopted model; and v) *Bayesian* solutions, such as Bayesian Patch Representation (BPR) [25], applying Bayes' theorem to extract features and using a probabilistic measure of similarity.

Finally, the *hand-crafted based* face recognition solutions derive features by extracting elementary, *a priori* selected, characteristics of the visual information. Usually these solutions are not very sensitive to face variations, e.g. pose, expression, occlusion, aging, and illumination changes, as they can consider multiple scales, orientations, and frequency bands. These solutions typically require tuning one or more parameters such as region size, scale, and topology. However, they are not computationally expensive in the enrolment phase as there is no need for training at feature extraction level. The hand-crafted based feature extraction approaches include: i) *shape based* solutions, such as Local Shape Map (LSM) [26], defining feature vectors using local shape descriptors; ii) *texture based* solutions, such as Local Binary Patterns (LBP) [27], exploring the structure of local spatial neighbourhoods; and ii) *frequency based* solutions, such as Local Phase Quantization (LPQ) [28], exploring the local structure in the frequency domain.

Moreover it is not uncommon to find hybrid face recognition solutions, such as LBP Net [29] or Mesh-LBP [30], combining elements from two or more feature extraction approaches to further improve the face recognition performance. Additionally, for face recognition solutions combining multiple features, extracted using different feature extraction methods, fusion can be done at several levels [31] [32]: i) *feature level*, usually concatenating features obtained by different feature extractors into a single vector for classification; ii) *score level*, combining the different classifier output scores, often using the 'sum rule'; iii) *rank level*, combining the ranking of the enrolled identities to consolidate the ranks output from multiple biometric systems; and iv) *decision level*, combining different decisions by those biometric matchers which provide access only to the final recognition decision, usually adopting a 'majority vote' scheme. Fusion at feature and score levels are the most commonly used approaches in the biometric literature. Generally, feature level fusion considers more information than score level fusion; however, it is not always possible to follow this method due to: i) incompatibility of features extracted in different feature spaces, notably in terms of data precision, scale, structure, and size; and ii) large dimensionality of the concatenated features, thus leading to a higher complexity in the matching stage [31]. If one of these difficulties exists, fusion can always be performed at the rank, score or decision levels.

## 4. Face recognition: *status quo*

This section exercises the proposed multi-level face recognition taxonomy by surveying the most representative available face recognition solutions under its umbrella. Table 2 summarizes the main characteristics of a selection of representative face recognition solutions, sorted according to the adopted feature extraction approach and sub-approach and their publication dates. In addition to the classes associated to the proposed taxonomy levels, this table also includes information about the face databases considered by these solutions for performance assessment purposes.

**Table 2:** Classification of a selection of representative face recognition solutions based on the proposed taxonomy.
*The abbreviations used in this table are defined in the footnote[1].*

| Solution Name/Acronym | Year | Face Structure | Feature Support | Feature Extraction Approach | Feature Extraction Sub-Approach | Database |
|---|---|---|---|---|---|---|
| PCA [15] | 1991 | Global | Global | Appearance | Linear | Private |
| ICA [16] | 2002 | Global | Global | Appearance | Linear | FERET |
| ASVDF [33] | 2016 | Global | Global | Appearance | Linear | PIE; FEI; FERET |
| KPCA MM [17] | 2016 | Global | Global | Appearance | Non-Linear | Yale; ORL |
| AHFSVD-Face [34] | 2017 | Global | Global | Appearance | Non-Linear | CMU PIE; LFW |
| GPCA [18] | 2004 | Global | Global | Appearance | Multi-Linear | AR; ORL |
| MPCA Tensor [35] | 2008 | Global | Global | Appearance | Multi-Linear | N/A |
| EBGM [19] | 1997 | Comp.+ Struct. | Global | Model | Graph | FERET |
| Homography Based [36] | 2017 | Comp.+ Struct. | Global Local | Model | Graph | FERET; CMU-PIE; Multi-PIE |
| 3DMM [20] | 2003 | Comp.+ Struct. | Global | Model | Shape | CMU-PIE; FERET |
| U-3DMM [37] | 2016 | Comp.+ Struct. | Global | Model | Shape | Multi-PIE; AR |
| Face Hallucination [38] | 2016 | Comp.+ Struct. | Global | Learning | Dictionary Learning | Yale B |
| Orthonormal Dic.. [39] | 2016 | Global | Global | Learning | Dictionary Learning | AR |
| LKED [22] | 2017 | Global | Global | Learning | Dictionary Learning | AR; FERET; CAS-PEAL |
| Decision Pyramid [23] | 2017 | Global | Local | Learning | Decision Tree | AR; Yale B |
| Logistic Regression [24] | 2014 | Global | Global | Learning | Regression | ORL; Yale B |
| BPR [25] | 2016 | Global | Local | Learning | Bayesian | AR |
| AlexNet [40] | 2014 | Global | Global | Learning | Deep Neural Net. | LFW; YTF |
| VGG Face [21] | 2015 | Global | Global | Learning | Deep Neural Net. | LFW; YTF |
| GoogLeNet [41] | 2015 | Global | Global | Learning | Deep Neural Net. | LFW |
| TRIVET [42] | 2016 | Global | Global | Learning | Deep Neural Net. | CASIA |
| Deep HFR [43] | 2016 | Global | Global | Learning | Deep Neural Net. | CASIA |
| Deep RGB-D [44] | 2016 | Global | Global | Learning | Deep Neural Net. | Kinect Face DB |
| CDL [45] | 2017 | Global | Global | Learning | Deep Neural Net. | CASIA |
| Deep NIR-VIS [46] | 2017 | Global | Global | Learning | Deep Neural Net. | CASIA |
| Deep CSH [47] | 2017 | Global | Global | Learning | Deep Neural Net. | CASIA |
| Lightened CNN [48] | 2018 | Global | Global | Learning | Deep Neural Net. | LFW; YTF |
| SqueezeNet [49] | 2018 | Global | Global | Learning | Deep Neural Net. | LFW |
| CCL ResNet [50] | 2018 | Global | Global | Learning | Deep Neural Net. | LFW |
| Cosface ResNet [51] | 2018 | Global | Global | Learning | Deep Neural Net. | LFW |
| Arcface ResNet [52] | 2018 | Global | Global | Learning | Deep Neural Net. | LFW |
| Ring loss ResNet [53] | 2018 | Global | Global | Learning | Deep Neural Net. | LFW |
| VGG-D[3] [54] | 2018 | Global | Global | Learning | Deep Neural Net. | LLFFD |
| VGG+ ConvLSTM [55] | 2018 | Global | Global | Learning | Deep Neural Net. | LLFFD |
| VGG+ LF-LSTM [56] | 2018 | Global | Global | Learning | Deep Neural Net. | LLFFD |
| LSM [26] | 2004 | Global | Local | Hand-Crafted | Shape | Private |
| LBP [27] | 2006 | Global | Local | Hand-Crafted | Texture | FERET |
| HOG [57] | 2011 | Global | Local | Hand-Crafted | Texture | FERET |

[1] ALTP, Adaptive Local Ternary Pattern; ASVDF, Adaptive Singular Value Decomposition Face; AHFSVD, Adaptive High-Frequency Singular Value Decomposition; BPR, Bayesian Patch Representation; BU-3DFE, Binghamton University 3D Facial Expression; CASIA, Chinese Academy of Sciences, Institute of Automation; CCL, Centralized Coordinate Learning; CDL, Coupled Deep Learning; Conv-LSTM, Conventional Long Short-Term Memory; CSH, Cross-Spectral Hallucination CSLBP, Center-Symmetric Local Binary Patterns; DCP, Dual-Cross Pattern; DLBP, Depth Local Binary Pattern; DM, Depth Map; ELBP, Extended Local Binary Patterns; FERET, FacE REcognition Technology; GPCA, Generalized Principle Component Analysis; HFR, Heterogeneous Face Recognition; HOG, Histogram of Oriented Gradients; KED, Kernel Extended Dictionary; KPCA MM, Kernel Principle Component Analysis Mixture Model; LBPNET, Local Binary Pattern Network; LKED, Learning Kernel Extended Dictionary; LCCP, Local Contourlet Combined Patterns; LDF, Local Difference Feature; LF, Light Field; LFHG, Light Field Histogram of Gradients; LFLBP, Light Field Local Binary Patterns; LF-LSTM, Light Field Long Short-Term Memory; LFW, Labelled Faces in the Wild; LiFFID, Light Field Face and Iris Database; LLFFD, Lenslet Light Field Face Database; LSM, Local Shape Map, LSTM, Long Short-Term Memory; LPS, Local Pattern Selection; MB-LBP, Multi-scale Block Light Field Local Binary Patterns; MDML-DCP, Multi-Directional Multi-Level Dual-Cross Pattern; ME-CS-LDP, Multi-resolution Elongated Centre-Symmetric Local Derivative Pattern; MF, Multi-Face; MLBP, Multi-scale Local Binary Pattern; MPCA, Multilinear Principal Component Analysis; MSB, Multi-Scale Blocking; ; ORL, Olivetti Research Ltd PCANET, Principle Component Analysis Network; RBP, Riesz Binary Pattern; TRIVET, TransfeR Nir-Vis heterogeneous facE recognition neTwork; U-3DMM, Unified 3D Morphable Model; VGG, Visual Geometry Group; VGG-D3, 2D+Disparity+Depth VGG; WPCA, Weighted Principle Component Analysis; YTF, YouTube Faces.

| Solution Name/Acronym | Year | Face Structure | Feature Support | Feature Extraction Approach | Feature Extraction Sub-Approach | Database |
|---|---|---|---|---|---|---|
| DLBP [58] | 2014 | Global | Local | Hand-Crafted | Texture | TEXAS; FRGC;BOSPHOR |
| ELBP [59] | 2016 | Global | Local | Hand-Crafted | Texture | Yale; FERET; CAS-PEAL |
| MB-LBP [60] | 2016 | Global | Local | Hand-Crafted | Texture | Yale B; FERET |
| MR CS-LDP [61] | 2016 | Component | Local | Hand-Crafted | Texture | PIE; Yale B; VALID |
| ALTP [62] | 2016 | Global | Local | Hand-Crafted | Texture | FERET;ORL |
| Face-Iris MF LF [63] | 2016 | Global | Local | Hand-Crafted | Texture | LiFFID |
| DM LF [64] | 2016 | Global | Local | Hand-Crafted | Texture | Private |
| LFLBP [65] | 2017 | Global | Local | Hand-Crafted | Texture | LLFFD |
| LFHG [66] | 2018 | Global | Local | Hand-Crafted | Texture | LLFFD |
| LPQ [28] | 2008 | Global | Local | Hand-Crafted | Frequency | CMU PIE |
| Hybrid Solution: Mesh-LBP [30] | 2015 | Comp.+ struct; | Local Global | Hand-Crafted; Model based | Texture; Graph | MIT CSAIL; BU-3DFE |
| Hybrid Solution: LBP Net [29] | 2016 | Global | Local | Hand-Crafted; Learning | Texture; Deep Neural Net. | LFW; FERET |
| Hybrid Solution: PCA Net [67] | 2016 | Global | Global | Appearance; Learning | Linear; Deep Neural Net. | LFW |
| Hybrid Solution: Aging FR [68] | 2016 | Component | Local | Hand-Crafted; Learning | Texture; Decision tree | MORPH |
| Hybrid Solution: MSB LBP+WPCA [69] | 2016 | Global | Local Global | Hand-Crafted; Appearance | Texture; Linear | ORL |
| Hybrid Solution: LFD+PCA [70] | 2016 | Comp.+ struct. | Local Global | Hand-Crafted; Appearance | Texture; Linear | SGIDCDL; FERET |
| Hybrid Solution: DeepBelief+CSLBP [71] | 2016 | Global | Local Global | Hand-Crafted; Learning | Texture; Deep Neural Net. | ORL |
| Hybrid Solution: Discriminative Dic. [72] | 2016 | Global | Local Global | Local Global | Texture; Deep Neural Net. | AR; Yale B |
| Hybrid Solution: Nonlinear 3DMM [73] | 2018 | Comp.+ struct. | Global | Appearance; Model; Learn. | Non-Linear; Shape; Deep Neural Net. | FaceWarehouse |
| Fusion Scheme: RGB-D-T [74] | 2014 | Global | Local | Hand-Crafted | Texture | Private |
| Fusion Scheme:RBP [75] | 2016 | Global | Local | Hand-Crafted | Texture | AR; Yale B; UMIST |
| Fusion Scheme: LCCP [76] | 2016 | Global | Local | Hand-Crafted | Frequency; Texture; | FERET |
| Fusion Scheme: Gabor-Zernike Des. [77] | 2016 | Global | Local | Hand-Crafted | Texture | ORL; Yale; AR |
| Fusion Scheme: MDML-DCP [78] | 2016 | Comp.+ struct. | Local Global | Hand-Crafted; Appearance | Texture; Linear | FRGC; CAS; FERET |
| Fusion Scheme: RGB-D-NIR [79] | 2016 | Global | Local Global | Hand-Crafted; Learning | Texture; Deep Neural Net. | Private |
| Fusion Scheme: Thermal Fusion [80] | 2016 | Global | Local | Hand-Crafted | Texture | Thermal/Visible Face |

The face recognition solutions listed in Table 2 which were proposed after 2014, are also included in Figure 4 to better illustrate the recent technological trends. Each vertical line in Figure 4 corresponds to one face recognition solution, which is classified according to the four levels of the proposed taxonomy, and the publication year. The information about face structure, feature support and publication year are represented in the X, Y, and Z directions, respectively; moreover, the vertical lines' color and marker symbol corresponds to the feature extraction approach and sub-approach, respectively.

The analysis of Figure 4 allows reaching some interesting conclusions related to the recent evolution of face recognition technology based on the proposed taxonomy levels, notably:

- **Face structure**: *Component* is the less considered face structural representation class, corresponding to only 4% of the listed solutions. The *component+structure* representation has recently received more attention, mostly by model based face recognition solutions, corresponding to more than 15% of the recent face recognition solutions surveyed. *Global* face representation is the most adopted face structure class, corresponding to more than 80% of the recent surveyed solutions.

- **Feature support**: *Global* and *local* feature supports have been equally considered for developing recent face recognition solutions. The selection of the feature support mostly depends on the feature extraction approach considered and both classes seem to be relevant.

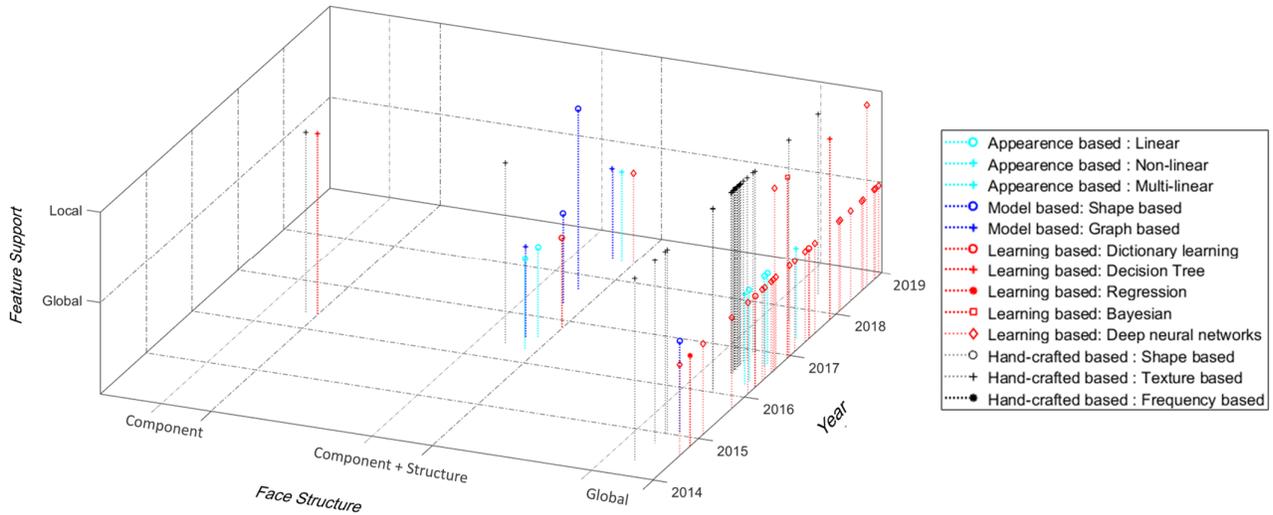

**Figure 4:** Visualization of a representative set of the last five years face recognition solutions, according to the four levels of the proposed taxonomy and publication date.

- **Feature extraction approach**: In recent years, learning based and hand-crafted based feature extraction approaches have gained increasing impact in the area of face recognition. Nowadays, thanks to the performance boosting brought by deep learning based solutions, learning based is the most popular feature extraction approach, as can be observed in Figure 4.

- **Feature extraction sub-approach**: From 2014 until 2016, texture hand-crafted based solutions were mostly proposed in the area of face recognition. Then deep learning based solutions gained momentum and are now dominating the face recognition state-of-the-art. In fact, more than 80% of the listed face recognition solutions proposed in 2018 were based on deep learning.

Additionally, the main trends on the evolution of face recognition solutions over a longer time frame are illustrated in Figure 5. Here, the recognition solutions are grouped based on their feature extraction approaches, and have associated the typical expected accuracy when tested on the LFW database [81]. Appearance based solutions dominated the face recognition landscape from the early 1990s until around 1997. Then, model based solutions emerged and remained the state-of-the-art until approximately 2006. After, hand-crafted based solutions gained momentum, further improving the accuracy of face recognition solutions. In 2014, DeepFace [40] dramatically boosted the state-of-the-art accuracy, from around 80% to above 97.5%. From that date, the face recognition research focus has shifted to deep learning based solutions and the current face recognition state-of-the-art is dominated by deep neural networks [82], offering near perfect accuracy for the LFW database.

A comprehensive evaluation of deep learning models for face recognition using different CNN architectures and different databases, under various facial variations, is available in [83]. Additionally, the impact of different covariates, such as compression artefacts, occlusions, noise, and color information, on the face recognition performance for different CNN architectures, including AlexNet [40], SqueezeNet [49], GoogLeNet [41], and VGG-Face [21], has been reported in [49]. The results have shown that the VGG-Face descriptor [21], computed using a VGG-16 network, achieves superior recognition performance under various facial variations, and is more robust to different covariates, when compared to relevant alternatives. More recently, deep learning based solutions based on the ResNet CNN architectures have been considered [50] [51] [52] [53], providing state-of-the-art results for face recognition by boosting the performance to above 99.8% for the LFW database.

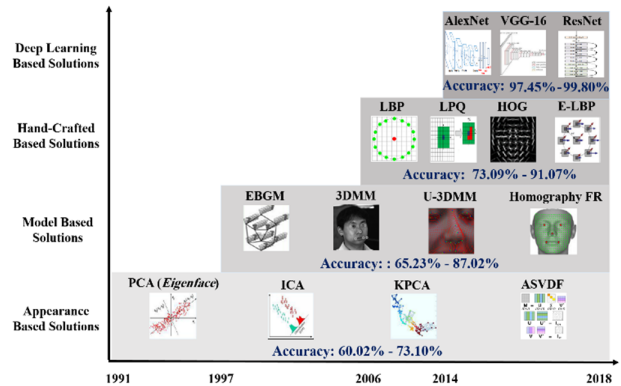

**Figure 5:** Evolution of face recognition solutions over time, grouped based on the feature extraction approach; indicative accuracy performance values for the LFW database.

## 5. Challenges and future research directions

This paper provides a survey of face recognition solutions based on a new, more encompassing and richer multi-level taxonomy for face recognition solutions. The proposed taxonomy facilitates the organization, categorization and comparison of face recognition solutions and may also guide researchers in the future development of more efficient face recognition solutions.

Based on the survey, it is possible to identify some face recognition algorithmic and application related challenges, which define future research directions, as follows.

**Algorithmic research challenges**

- **Double-deep face recognition for sequential biometric data** – Some emerging sensors may provide sequential biometric data, e.g., spatio-temporal or spatio-angular information. As most deep learning based face recognition solutions are designed to exploit the spatial information available in a 2D face image, developing double-deep solutions, combining convolutional neural networks and recurrent neural networks for jointly exploiting the available double information, has recently been considered [84] [55] [56]; this double approach is expected to be increasingly adopted as part of new biometric recognition solutions.

- **Face recognition for new sensors** - The emergence of novel imaging sensors, e.g., light field, NIR, Kinect, offers new and richer imaging modalities that can be exploited by face recognition systems [85]. As each imaging modality has its own specific characteristics, adopting face recognition solutions able to exploit the additional information available in the new imaging formats is a pressing need. For instance, the wide availability of dual- and multiple-cameras on mobile phones, as illustrated in Figure 6, will revolutionize the face recognition design to exploit the additional information provided by multiple cameras in mobile phones.

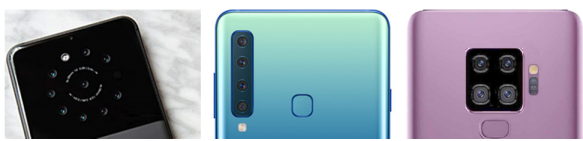

**Figure 6:** Mobile phones equipped with multiple-cameras [86] [87].

- **Deep face recognition model compression** – Deep face recognition solutions are nowadays dominating the face recognition arena; however, they are computationally complex. In this context, distillation and pruning techniques, exploring the deep model parameters redundancy to remove uncritical information, can be adopted to perform model compression and acceleration, without significantly reducing the recognition performance [95] [96]. This may facilitate the deployment of deep face models on resource-constrained embedded devices such as mobile devices [97].

- **Face presentation attack detection for face recognition security** - Despite the significant progresses in face recognition performance, the widespread adoption of face recognition solutions raises new security concerns. The security of a face recognition system can be compromised at different points, all the way from the biometric trait presentation to the final recognition decision. Face presentation attacks are among the most important security concerns nowadays, as they can performed by presenting face artefacts in front of the acquisition sensors, for instance using printed faces, electronic devices displaying a face or silicon face masks. Not to mention the possibility to generate fake faces using Generative Adversarial Networks (GANs) [91]. Thus, it is critical to incorporate in the face recognition systems efficient Presentation Attack Detection (PAD) solutions [92] [93] [94] in order security is not at risk.

- **Face identification from photo-sketches** - Photo-sketches are extremely important for forensics and law enforcement to identify suspects. A few face sketch recognition solutions have recently been proposed [88] [89] [90], relying on image-to-image translation for transforming a photo to a sketch and vice-versa. Sketch based face recognition is gaining more attention and becoming a hot topic, notably for situations where photos are not available.

**Application related challenges**

- **Face recognition in embedded systems** - Recent works have introduced specific hardware platforms and embedded systems to implement deep CNN architectures [98] [99]. The hardware implementation of deep face recognition systems in embedded systems is a critical step forward towards designing fast and accurate recognition solutions operating in real-time.

- **Face authentication in mobile phones** - Face authentication is becoming more and more popular on mobile phones, thus allowing mobile applications to verify the user's identity for granting access to sensitive mobile services such as e-banking. Different face recognition aspects, including presentation attacks and deployment constraints, should be investigated, notably in mobile environments [100] [101].

- **Face recognition in autonomous vehicles** - While a future with fully autonomous vehicles is easily predictable, security may be the key to its adoption. Face recognition systems can be integrated into the autonomous vehicles' existing systems to create a safer and more convenient driving experience. Face recognition systems can play a crucial role in the introduction of a new generation of keyless access control to vehicles, for unlocking doors and starting the engine. Additionally, recognition systems may personalize in-car settings, such as seat position and audio system settings, for each recognized driver or passenger [102].

- **Face analysis in health informatics** - Remarkable advances are being made in the area of health informatics due to the emergence of facial analysis systems (not necessarily face recognition systems). Facial analysis systems can be used in the context of pain diagnosis and management procedures such as detecting genetic diseases and tracking the usage of medication by a patient. The perspectives of facial analysis will critically shape health informatics services in the future [103].

- **Face recognition in smart cities** – According to a 2018 market research report [104], the biometrics market is estimated to grow from USD 13.89 billion in 2018 to reach USD 41.80 billion by 2023, where face recognition holds a substantial potential for growth during that period [105]. The face recognition market will have a huge potential in emerging application areas related to people identification in smart cities relying, notably electronic administration, smart houses, and smart education, among many others. [104].

Beyond the technical challenges, the use of public face recognition systems may bring ethical and privacy challenges and dilemmas related to the question of how the captured data, as a special category of personal data, may be used and manipulated. These issues start to be acknowledged by data protection regulations, such as EU General Data Protection Regulation (GDPR) [106], targeting the protection of people from having their information processed by a third party without their consent. In summary, there is a pressing need to protect biometric data by adopting new laws to regulate the usage of face recognition technologies [107] [108].

Face recognition technology is nowadays offering very efficient solutions for very different tasks and environments, ranging from forensics and surveillance to commerce and entertainment. This implies it is holding a huge potential for growth in the near and long term future and is going to be part of our everyday life, one way or another. This omnipresence brings not only natural technical challenges but also critical security, ethical and privacy challenges that should not be minimized as the way they will be addressed will largely determine the future of Human societies.